# Deep Learning at the Edge

Sahar Voghoei
*Department of Computer Science*
*University of Georgia*
Athens, GA, USA
voghoei@uga.edu

Navid Hashemi Tonekaboni
*Department of Computer Science*
*University of Georgia*
Athens, GA, USA
navidht@uga.edu

Jason G. Wallace
*Department of Crop & Soil Science*
*University of Georgia*
Athens, GA, USA
jason.wallace@uga.edu

Hamid R. Arabnia
*Department of Computer Science*
*University of Georgia*
Athens, GA, USA
hra@uga.edu

*Abstract*—The ever-increasing number of Internet of Things (IoT) devices has created a new computing paradigm, called edge computing, where most of the computations are performed at the edge devices, rather than on centralized servers. An edge device is an electronic device that provides connections to service providers and other edge devices; typically, such devices have limited resources. Since edge devices are resource-constrained, the task of launching algorithms, methods, and applications onto edge devices is considered to be a significant challenge. In this paper, we discuss one of the most widely used machine learning methods, namely, Deep Learning (DL) and offer a short survey on the recent approaches used to map DL onto the edge computing paradigm. We also provide relevant discussions about selected applications that would greatly benefit from DL at the edge.

## I. Introduction

The emergence of smart devices and sensors have tremendously expanded the scope of the Internet. Internet of things (IoT) refers to the ever-increasing network of objects and devices to collect and exchange data. Based on a report provided by Information Handling Services (IHS), there were about 15.4 billion worldwide connected devices in 2015, and it is expected to have more than 75 billion connected devices in 2025. The vast amount of data generated by these objects has motivated scientists to rethink about the conventional approaches to store and analyze the data. As a result, the concept of edge computing was proposed as a new computing paradigm in which the computation is mainly performed on the edge devices.

Edge computing resolves some of the critical challenges associated with IoT environments. For instance, in some IoT environments, such as healthcare, fast response is required. Rather than sending data to the cloud, edge computing minimizes the data transmission time and responds with low latency [1]. It also reduces the cost of data transmission, computation, and storage. Besides, processing data at the edge preserves the privacy of the users, since there is no need to upload the data to the cloud.

As a subclass of machine learning algorithms, Deep Learning (DL) has promised to provide a remarkable performance in many domains including image processing, speech recognition, and time-series data forecasting. Cameras, speakers, microphones, and multiple sensors are all located at the edge of the network which provides the great opportunity of running deep learning algorithms at the edge.

In this paper, we first discuss the challenges to have DL at the edge, then we provide promising approaches to address the challenges related to the resource limitation of the edge devices. Finally, the potentials and some applications of DL at the edge is presented.

## II. Challenges

In general, deep learning models consist of a large number of layers and parameters. For example, Deep Neural Network (DNN) and Convolutional Neural network (CNN) roughly use thousands of interconnected units and millions of parameters (sometimes over a billion) to train and use the models. As a result, the majority of mobile and other edge devices fail to train and run DL models in a timely manner.

There are multiple challenges associated with implementing and running deep learning algorithms on the edge devices. The totality of challenges pointed out in the literature of this field can be categorized as follows: resource-related issues, data-related issues, and security-related issues. Although the significance of the last two categories is not ignorable and may even result in redesigning the whole system [2,3], the primary objective of this paper is to provide a survey on the approaches that have been proposed to address the challenges associated with the limitedness of the resources at the edge.

The importance of the scope that will be covered in this paper is owed to the fact that limited computation, memory, and energy require new strategies to make a balance between the low latency which is provided at the edge, as opposed to the high-performance resources at the cloud. In other words, considering the inherent limitations of the edge devices, there is an increasing demand for the strategies to empower DL on them.

## III. Approaches

Edge devices, which are characterized by limited memory and storage, force developers to creatively commit to adopting various scale reduction techniques to shrink their usually huge deep learning networks [4]. In addition to the constraints mentioned above, limited computation performance, as well as limited power resources [5], are the other two essential characteristics of the edge devices. Consequently, optimal DL implementation and data/ model distribution techniques have been proposed to address the latter issues, respectively.

## A. DL Model Compression and Scale Reduction

Due to the limitations discussed earlier, even running a pre-trained DL model on edge devices will not be without difficulties [6]. Fortunately, some research studies found that many of the parameters used in Deep Neural Networks (DNNs) are redundant [8] and there is a high chance for most DL models to be over-parameterized. They suggest that a model can be efficiently simplified or compressed. Also, sometimes it is not necessary to use a very deep model, and it is readily feasible to eliminate some of the hidden layers without sacrificing a notable amount of accuracy [9]. As a result of removing these parameters (and layers), the complexity will be reduced [8], which makes the application suitable for the edge. We will discuss these approaches in more details below.

### 1) Pruning:

The concept of pruning consists in removing the unnecessary, less relevant, or sensitive (regarding privacy) links from the network, to create a smaller and less complicated model [10]. Pruning helps to reduce computation costs as well as storage and memory, while it retains the performance unaffected. Besides, pruning can improve the model by reducing the number of parameters, which generates more general models. An excellent way to identify redundant information is to rank the neurons by the size of their contribution to the result. Then, it will be easy to decide which links should be removed to shrink the network.

Traditionally, pruning has be done using manual thresholding, or hyperbolic and exponential biases [12], or second-order derivatives [13]. But, in more recent approaches in the context of mobile or edge application, there are some regularization-based approaches, like [15], which use regularization to reduce non-zero connections. The result of their approach on the AlexNet reduced the memory consumption by a factor of 4 without performance loss.

Structured Sparsity Learning (SSL), proposed by [16] uses regularization techniques to select a compact network from DNN and empower the structure by increasing its flexibility. Song Han [17] used L1 and L2 norms to prune the network and reduced the number of AlexNet's parameters by a factor of 9 and VGG-16 by a factor of 13, with no loss of accuracy. Reference [18] used L0 norm regularization for the pruning. Since L0 is non-differentiable, this paper proposed a way to find the weights that should be set to zero collectively. The result of the system can be structured with SGD. Another approach, called Fisher Pruning [19] uses greedy pruning combined with distilling the knowledge [20] to obtain the 10-time speedup, with the same performance, on CAT2000 dataset.

The resulting pruned model is the irregular and sparse network which needs a new parameter representation for loading and the arithmetic operations. Moreover, it requires a new suitable parallel technique as well. Reference [21] introduced structured sparsity technique, which uses map-wise, kernel-wise, and intra-kernel strided sparsity features. The achievement of this study lies in pruning the CIFAR-10 network by 70% with only 1% loss of accuracy and the ability to perform in parallel.

### 2) Quantization:

The primary objective of quantization is to use less memory. It also may reduce the demand for computation power and increase the speed. To reduce the required quantity of bits, quantization techniques change the presentation of each parameter from 32-bit float to 8-bit or less.

This section presents four levels of quantization that are the most frequently discussed ones in the literature of DL at the edge, namely initialized models, parameters, activation function, and structure.

On the level of the initialized model, the quantized transferred model can be used as a starting point [22] or as a teacher network in a knowledge distillation training. In this case, the model structure, all the weight calculations, and activation functions are based on the quantization approach.

Quantization on the second level, which is the most common one, is concerned with parameters and aims to quantize them. There are different quantization techniques like fixed-point quantization discussed in [23] and [24], or vector quantization [25], which achieve different compression ratios and accuracies. References [22] and [26] claimed in their paper that it is better to leave the first and the last layer unmodified because they are more sensitive to weight pruning and they form a tiny portion of the network compared to other layers. Another experimental study [22] approves this claim.

The third level is the activation function replacement. The easiest way is to replace it with a binary, but it weakens the accuracy. As a better alternative, in most of the cases, a clipping function with hardcoded values [22, 27] or clopping values per layer [26] create a better map with more accurate results. Based on the observation of [28] the combination of weight and activation works very well because activation quantization will not affect the accuracy while the weight format alteration leads to saving a good deal of memory. The simultaneous employment of both activation and weight quantization allows optimizing the tradeoff between memory saving and accuracy.

The last level is the structure, which can be the subject of alteration. There are some attempts to modify the structure of networks to apply binary quantized methods. Replacing FP32 convolution with multiple binary ones [26] is an example of these attempts. However, these approaches may scale up the network and have a reverse side-effect in terms of memory and storage. Another example of operation on this level is to optimize the number of the DL layers through group sparsity regularization [29].

### 3) Hashing:

The other way to apply compression in DL is hashing. The most common type of hashing technique is to use low-cost hash functions to randomly cluster connection weights into hash buckets which share the same parameter values [30,31]. This approach can help in the case of sparsity representation that is mostly produced by pruning or filtering. HashedNet [32] uses this hashing approach with the help of pruning techniques to have more memory saving. The testing result of HashedNet shows that the accuracy of this model is not dropping much.

Another method of hashing is Locality Sensitive Hashing (LSH) that maps similar data into the same bucket with high probability. A very novel hashing method used LSH for maximum inner product search (MIPS) to select nodes with highest activation efficiency [33]. The critical point is that there is an LSH hash table with every layer and it is used to shrink the network size to 5% of neurons usage by losing only 1% of the accuracy. As a result of having sparse gradient updates, the algorithm is suited for asynchronous and parallel training.

Different from LSH, which is data-independent and unsupervised, more general methods can incorporate semantic labels or relevant information to mitigate the semantic gap which significantly improves the hashing quality. Minimal Loss Hashing (MLH) [34] and Supervised Hashing with Kernels (KSH) [35] are well-known structured methods. KSH generates hash codes by minimizing the Hamming distances across similar pairs and maximizing the Hamming distances across different pairs. In the context of deep supervised methods, CNNH [36], DNNH [37], DHN [38] are commonly used primarily in the case of limited available resource environments like mobile and other edge devices.

Reference [39] introduced a data-dependent continuation hash method, called HashNet, which learns exactly binary hash codes from imbalanced similarity data which shows better performance on hashing weights and activations. They compared their proposed model with all supervised and unsupervised models prior to their paper on three datasets (ImageNet, NUS-WIDE, and MS COCO). HashNet considerably outperformed the previous approaches.

*4) Hybrid:*

Some researchers proposed hybrid approaches incorporating both software and hardware be to perform the compression more efficiently while having parallelism methods in mind. For example, [40] introduced Cambricon-X accelerator to deal with sparse DL. Another study [41] proposed an energy-efficient inference engine, called EIE, with the specialized hardware architecture which applies compression and handles weight sharing efficiently with the help of pruning. EIE is an array of processing elements, where each processing element stores a partition of the network in SRAM, rather than DRAM.

Reference [42] extend the TensorFlow framework by adding pruning to the network's connections during training. This framework has a mobile version for deep learning and easily can be used on edge devices. Another study [43] proposed a Runtime Neural Pruning (RNP) framework to prune the neural network dynamically. Since the ability of network is fully preserved, the balance point is readily adjustable according to the available resources. RNP can be applied to off-the-shelf network structures and reaches a better tradeoff between speed and accuracy

*B. Optimal DL Implementation*

In DL at the edge, most of the times the data is offloaded to the cloud to train the models and then the trained model will be placed on the end devices. The model can be optimized using the new data on the cloud and replace the old model on the edge devices. Although it seems that optimization techniques are needed only in the training phase, there is a need to create a smaller model which require less computation and memory with minimum loss in the accuracy. Below, we explained some of the ideas that optimize the DL models to be more suitable for edge environments.

*1) Mathematical Optimization:*

In this section, we cover the optimization techniques that enable DL at the edge. For instance, there are three variants of Gradient Descent: Batch, Stochastic, and Mini-batch. If the dataset does not fit into the memory, batch gradient descent works very slow and doesn't perform as a suitable algorithm for online data. Stochastic Gradient Descent (SGD) performs one update at a time and runs much faster in comparison with the Batch Gradient, but the objective function of SGD heavily fluctuates which converges at different minima each time. Also, SGDs are difficult to tune and parallelize; therefore, it is difficult to scale up a DL algorithm with SGDs. Mini-batch, which is more stable than the approaches mentioned earlier, reduces the frequency of updating the parameters and efficiently computes gradients employing matrix optimization.

Methods like Limited Memory BFGS (L-BFGS) and Conjugate Gradient (CG), which use second-order approximation and conjugacy information for optimization are more stable and able to boost the performance of SGDs. These methods are more suitable for edge environment because they can manage the memory consumption and computational processes much better. These two algorithms are highly used with CNN, DBN, LSTM, AE, and MLP in different applications at the edge. Reference [44] compared the performance of SGD, SGD line-search, CG minibatch, and L-BFGS minibatch. They conclude that CG with a small minibatch, and LBFGS in the second place, perform much better than the other SGDs; while, in the parallel fashion, LBFGS outperforms the other ones.

Momentum-accelerated SGD [45] is based on SGD and accelerates gradients' vectors to go in the right direction, which achieves a faster convergence than SGD. It introduced a way to learn from the last gradient for the next round of the updates and impressively succeeds to escape local minima. Another optimizer based on Gradient Descent is AdaGrad (Adaptive Gradient Algorithm) [46], which uses different learning rates for a different parameter. Root Mean Square Propagation (RMSProp) based on averaging the recent magnitudes of gradients, uses different rates for parameters, which makes it an appropriate choice to handle noisy datasets and online data.

Later, Adadelta [47], having added self-adaptive learning rates to AdaGrad, managed to reduce the number of parameters to learn. It provided a faster performance in comparison with the latter; however, it still couldn't handle sparse gradients.

Adam (Adaptive Moment Estimation) method [48] is very similar to AdaGrad and RMSProb. It retains all the aforementioned advantages of both the latter. It uses the first and the second moments of the gradient in the average (instead of only first as AdaGrad does). It also uses an exponential moving average of the gradients and square gradients which controls the decay of these parameters in the algorithm. Adam is well known as a fast convergence algorithm for sparse gradient and non-stationary problems. This algorithm suited well with CNN, LSTM, and GAN for many applications like mobile traffic forecasting [49], cellular traffic prediction [50], and mobile traffic super-resolution [51]. AdaMax [43] is an improvement on Adam since it uses Lp norm instead of L1 or L2, which makes it more stable.

Nadam (Nesterov-accelerated Adaptive Moment Estimation) [52] is another algorithm, which has some improvements on Adam. It is a combination of Adam and NAG (Nesterov Moment Estimation). It adds some constraints to the gradients; consequently, it has a faster convergence.

To address the problem of not converging at the optimal solution in Adam and other algorithms, AMSGrad [53] uses the maximum of the past squared gradients, rather than an exponential average, for updating parameters. Consequently, it provides better performance on the accuracy for small datasets; but for large data, there should be more studies to verify the author's claim for beating Adam.

*2) Approximation:*

Predictive uncertainty estimation is an essential key to understanding and improving decision accuracy for mobile and other devices at the edge. Employing computational resource reduction, DL researchers tried to use uncertainty estimation to make DL more suitable for the edge devices. Reference [54] introduced MCDrop as the first attempt to use the linkage of dropout training and deep Gaussian process. The other attempt is SSP by [55], which is based on scoring rules and ensemble methods. Most of the uncertainty methods based on Bayesian Neural Networks (NN) are computationally expensive, and the ones based on the methods like sampling [54] and ensemble [55] are time-consuming.

More recent studies, such as RDeepSense [56], employ effective dropout training that interprets NNs as a Gaussian through Bayesian approximation and predictive estimation. It reduces the computation complexity. This method uses the scoring rule to smooth the underestimation effect of MCDrop and uses dropout to avoid the computational complexity of SSP.

## C. Data and Model Distribution

Talking about deep learning on the edge devices and their limitation, it may not be possible to fit a model on one device. On the other hand, most of the data created or gathered from the end devices are stored in a very distributed fashion, and it might not be reasonable to send them to a central server to be analyzed. The primary challenges for different distribution techniques on small and geographically distributed devices are the memory limitation, the power consumption and the computational capacity of devices. In general, there are two types of distribution: model parallelism or data parallelism.

*a) Model Parallelism:* is trying to distribute the memory and computational requirements by distributing the model itself, but due to the synchronization overhead, it is much slower than centralized approaches. Reference [46] introduced Downpur SGD (asynchronous stochastic gradient), and Sandblaster L-BFGS (distributed implementation of L-BFGS) technique. They combined these two techniques together to create a DistBelif framework which made it possible to train much bigger DL models (30 times bigger networks at the time of their study) using 16000 CPU cores on 1000 machines. Adapting this method to the more dynamic environment and using GPUs, [57] had success with their proposed Commodity Off-The-Shelf High-Performance Computing (COTS HPC) technique. They used GPUs and MPI to train up to 11 billion parameters just on 16 machines to show how scalable their approach is. The good news about their proposed method is that with a little bit of change it will be a very promising technique for edge environments. More recently, [58, 59] introduced a hierarchically distributed method from cloud to end devices, using local aggregators and binary weights to reduce both computational storage and communication overhead. Reference [60] used the small network on the local devices, and if they fail to do the classification job locally, then the larger DL model in the cloud will try to train based on all the data coming from devices. This architecture maintains good accuracy as well as considerable latency drop.

*b) Data parallelism:* this technique, which is also called federated learning, is distributing data between workers and collaboratively learn a shared model. There are many effective ways to distribute data like using a stereo-correlation algorithm which adopts a data farming approach to balance the workload [63, 64] or designing DL for learning the correlation between data and distributing it [7]. After each worker gets a partition of data, they perform the training on their data independently. Then, all the nodes should synchronize their parameters together. This way the whole process will speed up [61,62] while using less memory and computation power. This is the same techniques adopted by the DeepBelief model provided by Google [65]. There are three approaches talking about how workers synchronize their parameters to make an optimal model: Bulk Synchronous Parallel (BSP) [66], which syncs all updates before each iteration, Stale Synchronous Parallel (SSP) [67], which allows the faster workers to be some limited number of iteration ahead of the slower ones, and Total Asynchronous Parallel (TAP) [68], in which workers can be asynchronous. The latter method cannot guarantee to converge; therefore, it is used in very specific situations while the other two methods are commonly used in the edge environments.

Some frameworks are trying to use advance distribution techniques all together to improve the accuracy and the latency with minimum computation and communication cost. For instance, Adam [69] is very efficient and scalable architecture which exploits asynchrony throughout the system to improve the performance and accuracy (with 30x fewer devices and 2x higher accuracy). Also, their technique handled the delay and fault tolerance incredibly. As another example, GeePS [70] uses GPU to reach fast convergence rate and high training throughput suitable for DL distribution at the edge.

## IV. APPLICATIONS

The proliferation of mobile and IoT devices have introduced numerous applications where the immediate action upon sensing different features is essential. There are various domains such as smart cities, agriculture, healthcare, education, sports, and energy management [76,77,78,79] where various applications would benefit from IoT-based technologies. One of the well-known examples is self-driving cars in which the controlling devices have to perform appropriate actions with minimum latency. Virtual and augmented reality is another application where activity recognition and image classification at the edge could be used for entertainment and advertisement. Although there are many challenges associated with implementing DL at the edge devices [71], DNNs proved to be very promising in many fields such as object recognition and speech recognition. In this section, we point out some selected applications where local processing and prompt response is of great importance.

There have been an increasing number of applications for the pattern and objected recognition in the images. In autonomous vehicles, immediate detection of different objects and obstacles is required so that the controllers can make a decisive action to change the speed, activate the braking system, or change the route. For instance, [72] proposed a model to take the RGB images along with the LIDAR point cloud images to predict the 3D bounding box for the objects. Their proposed deep fusion network provides the opportunity for autonomous cars to achieve higher performance and increase their safety. As another example, YOU Only Look Once (Fast YOLO) [73] application that is an energy efficient close-to-real-time image classification application that can classify 155 frames per second (fps) images and is one of the DL frameworks which works on most mobile devices. As another example, deep learning algorithms on surveillance cameras can promptly detect human activities and send appropriate notifications. Reference [74], proposed a simple deep learning network to analyze the frames in a video sequence to detect fall detection. Other than health-related applications, detecting the unusual behavior of people is another potential application which could be addressed using the surveillance cameras.

Deep neural networks have also been performed very well in the domain of speech recognition. Speech recognition has a high potential in wearable and other pervasive devices where they can enable people to have a more efficient collaboration with each other and also provides the opportunity to have easier access to the information. For instance, smart speakers and virtual assistants such as Google Home and Amazon Alexa have become very popular in recent years. With the emergence of deep learning and edge computing technologies, there have been many research studies trying to incorporate DL-based speech recognition approaches into the edge devices for limited vocabulary speech recognition applications. Reference [75] studied the human-machine collaboration design to build speech recognition systems based on DNN architecture. They proposed a family of DNNs, called EdgeSpeechNets, suitable for the resource-constrained edge devices. Their experimental results on the Google Speech Commands dataset could reach around 97% accuracy, with a much smaller network (around eight times smaller than the original network). Instant translation systems are another subcategory of speech recognition which many companies such as Skype have been working on.

## V. CONCLUSION

This manuscript is not meant to be a detailed and elaborate survey paper. It only provides a short and concise survey on solutions proposed to address the challenges associated with having DL at the edge. A conscious attempt was made to consider only the recent advances in this important and growing area. We provided our perspective and discussed the potential of growth and need for DL at the edge.

International Internet of Things Summit, pp. 484-492. Springer, Cham, 2015.

[61] McMahan, H. Brendan, Eider Moore, Daniel Ramage, and Seth Hampson. "Communication-efficient learning of deep networks from decentralized data." arXiv preprint arXiv:1602.05629 (2016).

[62] McMahan, Brendan, and Daniel Ramage. "Federated learning: Collaborative machine learning without centralized training data." Google Research Blog (2017).

[63] Arabnia, Hamid R. "Distributed stereo-correlation algorithm." Computer Communications 19, no. 8 (1996): 707-711.

[64] Wani, M. Arif, and Hamid R. Arabnia. "Parallel edge-region-based segmentation algorithm targeted at a reconfigurable multiring network." The Journal of Supercomputing 25, no. 1 (2003): 43-62.

[65] Girija, Sanjay Surendranath. "Tensorflow: Large-scale machine learning on heterogeneous distributed systems." (2016).

[66] Valiant, Leslie G. "A bridging model for parallel computation." Communications of the ACM 33, no. 8 (1990): 103-111.

[67] Ho, Qirong, James Cipar, Henggang Cui, Seunghak Lee, Jin Kyu Kim, Phillip B. Gibbons, Garth A. Gibson, Greg Ganger, and Eric P. Xing. "More effective distributed ml via a stale synchronous parallel parameter server." In Advances in neural information processing systems, pp. 1223-1231. 2013.

[68] Recht, Benjamin, Christopher Re, Stephen Wright, and Feng Niu. "Hogwild: A lock-free approach to parallelizing stochastic gradient descent." In Advances in neural information processing systems, pp. 693-701. 2011.

[69] Chilimbi, Trishul M., Yutaka Suzue, Johnson Apacible, and Karthik Kalyanaraman. "Project Adam: Building an Efficient and Scalable Deep Learning Training System." In OSDI, vol. 14, pp. 571-582. 2014.

[70] Cui, Henggang, Hao Zhang, Gregory R. Ganger, Phillip B. Gibbons, and Eric P. Xing. "GeePS: Scalable deep learning on distributed GPUs with a GPU-specialized parameter server." In Proceedings of the Eleventh European Conference on Computer Systems, p. 4. ACM, 2016.

[71] Rallapalli, S., H. Qiu, A. Bency, S. Karthikeyan, R. Govindan, B. Manjunath, and R. Urgaonkar. "Are very deep neural networks feasible on mobile devices." IEEE Trans. Circ. Syst. Video Technol(2016).

[72] Yao, Shuochao, Yiran Zhao, Huajie Shao, Aston Zhang, Chao Zhang, Shen Li, and Tarek Abdelzaher. "Rdeepsense: Reliable deep mobile computing models with uncertainty estimations." Proceedings of the ACM on Interactive, Mobile, Wearable and Ubiquitous Technologies 1, no. 4 (2018): 173.

[73] Redmon, Joseph, Santosh Divvala, Ross Girshick, and Ali Farhadi. "You only look once: Unified, real-time object detection." In Proceedings of the IEEE conference on computer vision and pattern recognition, pp. 779-788. 2016.

[74] Chen, Xiaozhi, Huimin Ma, Ji Wan, Bo Li, and Tian Xia. "Multi-view 3d object detection network for autonomous driving." In IEEE CVPR, vol. 1, no. 2, p. 3. 2017.

[75] Gal, Yarin, and Zoubin Ghahramani. "Dropout as a Bayesian approximation: Representing model uncertainty in deep learning." In international conference on machine learning, pp. 1050-1059. 2016.

[76] Zhang, Chaoyun, Paul Patras, and Hamed Haddadi. "Deep Learning in Mobile and Wireless Networking: A Survey." arXiv preprint arXiv:1803.04311 (2018).

[77] Hashemi Tonekaboni, Navid, Sujeet Kulkarni, and Lakshmish Ramaswamy. "Edge-Based Anomalous Sensor Placement Detection for Participatory Sensing of Urban Heat Islands" In Smart Cities Conference (ISC2), 2018 International, pp. 1-8. IEEE, 2018

[78] Mohammadi, Mehdi, Ala Al-Fuqaha, Sameh Sorour, and Mohsen Guizani. "Deep Learning for IoT Big Data and Streaming Analytics: A Survey." IEEE Communications Surveys & Tutorials (2018).

[79] Ota, Kaoru, Minh Son Dao, Vasileios Mezaris, and Francesco GB De Natale. "Deep learning for mobile multimedia: A survey." ACM Transactions on Multimedia Computing, Communications, and Applications (TOMM) 13, no. 3s (2017): 34.